\documentclass{article} 
\usepackage[preprint]{aaai2026}  
\usepackage{times}  
\usepackage{helvet}  
\usepackage{courier}  
\usepackage[hyphens]{url}  
\usepackage{graphicx} 
\usepackage{natbib}  
\usepackage{caption} 
\usepackage{algorithm}
\usepackage{algorithmic}

\usepackage{amsmath}
\usepackage{multirow}
\usepackage{booktabs}
\usepackage{makecell}
\usepackage{tikz}
\usepackage{colortbl}
\usepackage{amssymb}
\usepackage{pifont}
\usepackage{subcaption}
\usetikzlibrary{shapes,shadows}
\usepackage[most]{tcolorbox}
\usepackage{tabularx}
\definecolor{my_green}{RGB}{50,220,0}
\definecolor{my_yellow}{RGB}{255,165,0}
\definecolor{my_red}{RGB}{255, 0, 0}
\definecolor{my_black}{RGB}{0, 0, 0}
\definecolor{error_red}{RGB}{241,156,153}
\definecolor{pass_green}{RGB}{185,224,165}
\definecolor{shadecolor}{RGB}{237,237,237}
\newcommand{\cmark}{\textcolor{my_green}{\ding{51}}} 
\newcommand{\xmark}{\textcolor{my_red}{\ding{55}}} 

\usepackage{newfloat}
\usepackage{listings}
\DeclareCaptionStyle{ruled}{labelfont=normalfont,labelsep=colon,strut=off} 
\floatstyle{ruled}
\newfloat{listing}{tb}{lst}{}
\floatname{listing}{Listing}
\title{P-Aligner: Enabling Pre-Alignment of Language Models\\via Principled Instruction Synthesis}
\author {
    Feifan Song,
    Bofei Gao,
    Yifan Song,
    Yi Liu,
    Weimin Xiong,
    Yuyang Song\\
    Tianyu Liu,
    Guoyin Wang,
    Houfeng Wang
}
\affiliations {
    State Key Laboratory of Multimedia Information Processing, School of Computer Science, Peking University\\
    songff@stu.pku.edu.cn; wanghf@pku.edu.cn
}

\usepackage{bibentry}

\begin{document}

\maketitle

\begin{abstract}
Large Language Models (LLMs) are expected to produce safe, helpful, and honest content during interaction with human users, but they frequently fail to align with such values when given flawed instructions, e.g., missing context, ambiguous directives, or inappropriate tone, leaving substantial room for improvement along multiple dimensions.
A cost-effective yet high-impact way is to pre-align instructions before the model begins decoding.
Existing approaches either rely on prohibitive test-time search costs or end-to-end model rewrite, which is powered by a customized training corpus with unclear objectives.
In this work, we demonstrate that the goal of efficient and effective preference alignment can be achieved by \textbf{P-Aligner}, a lightweight module generating instructions that preserve the original intents while being expressed in a more human-preferred form.
P-Aligner is trained on \textbf{UltraPrompt}, a new dataset synthesized via a proposed principle-guided pipeline using Monte-Carlo Tree Search, which systematically explores the space of candidate instructions that are closely tied to human preference.
Experiments across different methods show that P-Aligner generally outperforms strong baselines across various models and benchmarks, including average win-rate gains of 28.35\% and 8.69\% on GPT-4-turbo and Gemma-2-SimPO, respectively.
Further analyses validate its effectiveness and efficiency through multiple perspectives, including data quality, search strategies, iterative deployment, and time overhead.
\end{abstract}

\begin{links}
    \link{Code}{github.com/F2-Song/P-Aligner}
    \link{Data}{huggingface.co/datasets/songff/UltraPrompt}
\end{links}

\section{Introduction}
Large Language Models (LLMs) have demonstrated remarkable versatility across a broad range of downstream tasks without any task-specific fine-tuning~\citep{brown2020language, achiam2023gpt}, and the strategic design of prompts has proven to further amplify their effectiveness~\citep{wang2023promptagent, pryzant2023automatic}.
Nevertheless, when the objective shifts to aligning outputs with human preferences, e.g. expectations in harmlessness, helpfulness and honesty, even extensively post-trained models remain fragile, indicating that effective preference learning is still an open challenge.
While some failures can be attributed to limitations in training data coverage, a non-negligible portion arises from the user inputs themselves: models are highly sensitive to subtle cues in insturctions.
Well-crafted instructions, like those with sufficient background, clear instructions, or an appropriate tone, greatly enhance the likelihood of obtaining satisfactory responses.
In contrast, flawed instructions often result in undesired outputs.

To pre-process instructions between the user input and LLM inference can be useful~\citep{li2024dialogue, cheng-etal-2024-black}.
For example, \citet{cheng-etal-2024-black} released a module for automatic instruction enhancement, which is trained on a corpus of refined instruction synthesized through a heuristic way.
However, it does not explicitly define how instructions should be polished, resulting in unclear improvement directions and limited performance gains.
This gives rise to a key challenge: how to design the method of instruction synthesis that explicitly searches for better instructions, thereby enhancing the quality of LLM-generated outputs?

\begin{figure}[t]
    \centering
    \includegraphics[width=0.98\linewidth]{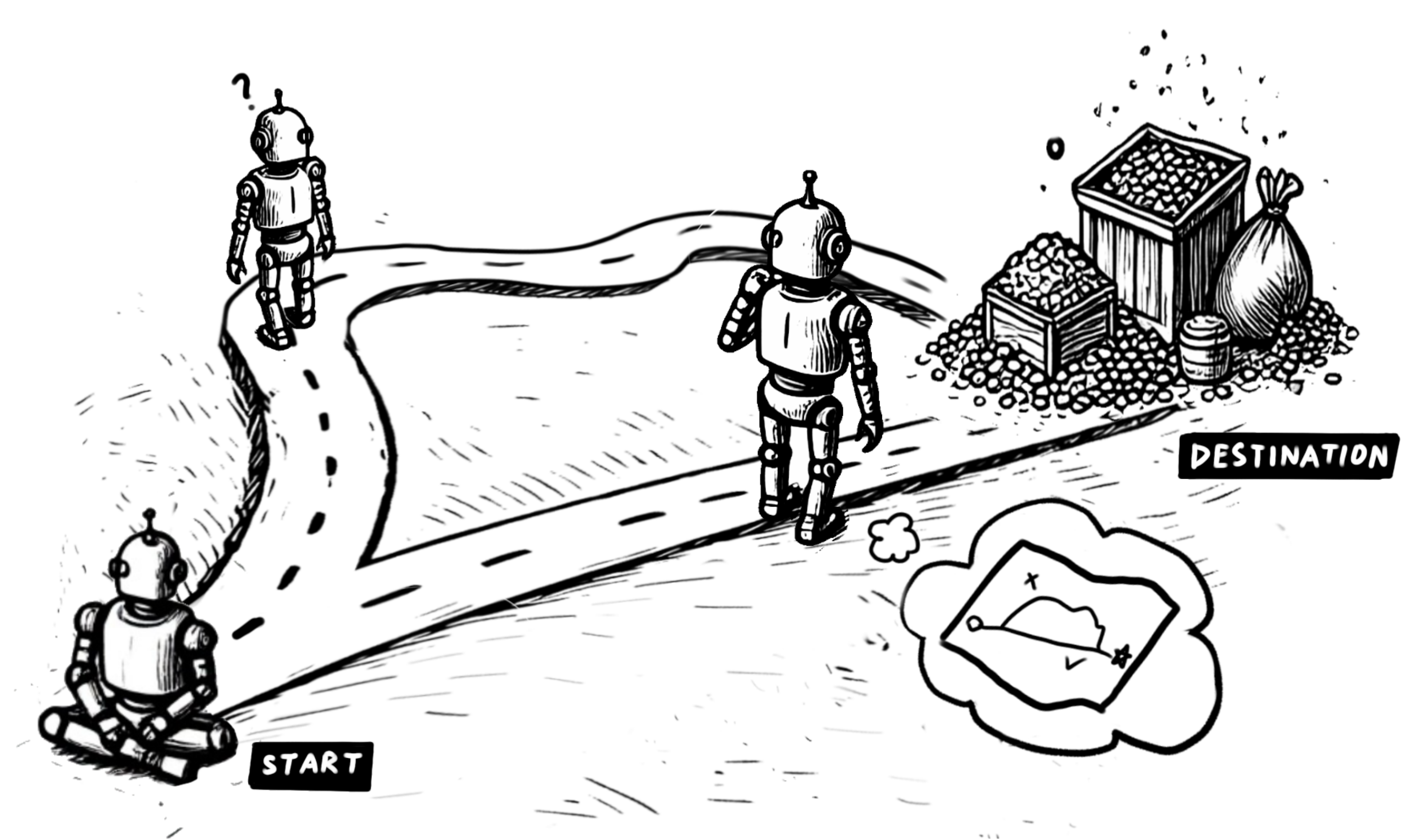} 
    \caption{
        Illustration of our motivation. Since instructions can leave room for improvement, feeding LLM raw instructions works like walking from the very origin (\emph{lower left bot}); Implicit rewrites can be helpful, yet lacking explicit supervision can cause a long path (\emph{upper bot}); 
        By contrast, our pipeline supplies an explicit trajectory that targets human preference with iterative search, yielding both higher alignment and lower overall steps (\emph{central bot}).
    }
    \label{fig:intro}
\end{figure}

To meet the above challenge, we propose to anchor the instruction-refinement process to a concrete principle set.
Each principle encodes an explicit direction, thereby converting the vague goal of \textbf{better} instructions into a finite, interpretable action space.
However, solely enumerating edit directions does not guarantee monotonic improvement and a quantitative feedback signal is also required, while directly scoring an instruction is infeasible at scale. To address it, we treat the quality of LLM responses as a measurable proxy: first sampling multiple continuations for a given instruction, then applying an off-the-shelf preference reward model to score the responses, which serve as a reliable signal for the instruction without human annotation.

With these two components, we design a new pipeline for synthesizing high-quality instructions that treats instruction refinement as iterative self-editing with pre-defined principles corresponing to general human preference, which is implemented by Monte Carlo Tree Search (MCTS). 
Each transformation is rewriting the last instruction state, regulated by one principle, thus incrementally improving the input through multi-step reasoning. 
Unlike other multi-step reasoning tasks, each intermediate instruction is also a potential output, and we accordingly customize Simulation, Backpropagation and other stages to accommodate it.

Building on this pipeline, we propose \textbf{P-Aligner}, a lightweight module between the user input and LLM inference, similar to \citet{cheng-etal-2024-black}.
Following the spirit of instruction synthesis, it automatically produces principled instructions end-to-end to empower LLM preference alignment. 
Specifically, we construct \textbf{UltraPrompt}, an instruction set with preference from top- and bottom-ranked instructions filtered out from each search tree.
P-Aligner demonstrates strong benefits to LLM inference while significantly reducing the time overhead.

We conduct comprehensive evaluations across competitive baselines and language models to evaluate its effectiveness, where P-Aligner raises the average win rate by 28.35\% over the best baseline on GPT-4-turbo and by 8.69\% on Gemma-2-SimPO, demonstrating consistent gains.
Ablations that control the data quality confirm that the synthesized corpus is the primary driver of improvement for P-Aligner, and benchmarking it against on-the-fly search implementations also shows its comparable performance at a fraction of the cost.
Finally, we analyze iterative application and real-world serving overhead, revealing that its one-shot optimization can approach the performance ceiling and incurs negligible latency under batched deployment.

Another gain of this work is \textbf{SinglePO}, a single-step principle-oriented rewriter acquired from UltraPrompt, which will be released along with P-Aligner to facilitate local and low-resource deployment of the data pipeline or further research by the community.

We summarize our contributions as follows:
\begin{itemize}
    \item[(1)]We design a new instruction synthesis pipeline with principles, allowing further training with controllable and interpretable synthesized data.
    \item[(2)]We build UltraPrompt, a synthesized instruction set from the pipeline, and gain P-Aligner, a lightweight module for LLM pre-alignment before inference.
    \item[(3)]We conduct comprehensive experiments and analysis to validate our approach, as well as propose another module, SinglePO, to support local low-resource deployment of the synthesis pipeline.
\end{itemize}

\section{Related Works}
\subsection{Preference Learning}
Despite extensive pre-training, large language models are not ready for direct human interaction for they risk producing undesired content, necessitating alignment with human preferences.
The dominant paradigm of preference learning is post-training on high-quality preference data, which shifts the token distribution at decoding time toward human-aligned outputs.
This approach has been instantiated via both reinforcement-learning~\citep{ouyang2022training, ji2024pku} and supervised-learning techniques~\citep{rafailov2024direct, meng2024simpo}.
However, the requirements of substantial data and computations remain challenging for most developers, encouraging researchers explore low-cost alternatives.
A popular line of such works concentrates on steering the distribution with external reward models~\citep{mudgal2024controlled, khanov2024args, li2024cascade}, in-context learning~\citep{lin2024the, li2024rain}, or directly rephrasing the response~\citep{ji2024aligner}.
Orthogonal to these efforts, optimization at the input side has also been proved effective. For example, \citet{cheng-etal-2024-black} trains an end-to-end module that rewrites user instructions, thereby guiding LLMs to produce safer and more helpful responses. Combining both two sides can further enhance preference leraning, as shown in our experiments.

\subsection{Prompt Engineering}
Prompt design has been the critical factor for activating and boosting potential performance on downstream tasks in the era of large language models~\cite{shin2020autoprompt}. To reduce the burden of human-crafted prompt design and further promote its effect, a batch of works propose different workflows that automatically re-construct or evolve user input.

\citet{zhou2022large} and \citet{wang2023promptagent} leverage Monte-Carlo Tree Search (MCTS) to iteratively explore candidate prompts, treating it as a sequential decision-making problem.
\citet{pryzant2023automatic} frames prompt revision as gradient descent on discrete text: at each step, the current prompt is tagged with feedback for next-time optimization.
\citet{li2023guiding} augments the user query with task-oriented hints, steering LLMs toward better completions.
Going beyond single-turn hints, \citet{wang2024minstrel} introduces structural context to organizing relevant background, constraints, and user intent into a coherent scaffold. 
However, above approaches are primarily evaluated on classical NLP benchmarks and lack principled designs for human-preference alignment.
Moreover, their reliance on test-time iterative optimization inflates overall response latency.
\citet{cheng-etal-2024-black} address both limitations with BPO, a lightweight rewriter that end-to-end map any user instruction to a preference-aligned reformulation, while its training corpus is constructed heuristically, limiting both coverage and quality.
We build upon these insights and propose a new synthesis pipeline explicitly tailored for human-preference alignment, which yields larger and higher-quality samples to train better module.

\begin{figure*}[t]
    \centering
    \includegraphics[width=\linewidth]{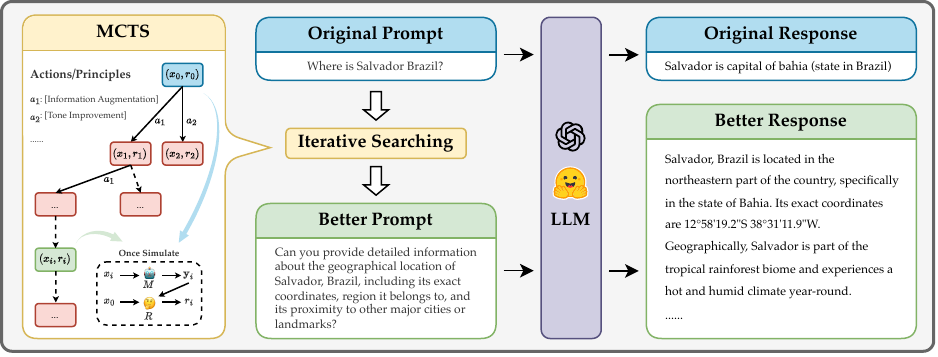} 
    \caption{Illustration of the proposed pipeline for principled instruction synthesis, based on MCTS. With iterative search, the produced better instruction can guide the LLM to return better responses.}
    \label{fig:head}
\end{figure*}

\section{Preliminary}

We begin by formalizing the task addressed in this work.  
Let $\mathcal{X}$ denote the space of user instructions and $\mathcal{Y}$ the space of LLM responses.  
The prevalent LLMs are designed to map an instruction \(x\in\mathcal{X}\) to a response \(y\in\mathcal{Y}\) that aligns with human preference,
\begin{equation}
    y \sim M(x)
\end{equation}
where $M$ represents an LLM. Such preference is commonly summarized by the \textbf{3H} criteria: \textbf{Helpfulness}, \textbf{Harmlessness}, and \textbf{Honesty}. 
However, $M$ can still fails to sample satisfactory responses, which can be attributed to either (i) its imperfect instruction-following ability, or (ii) the ambiguity, disrespectful tone, or other features of the instruction \(x\) that already violates the 3H criteria and misguides the LLM prior to generation. 

To mitigate the potentially negative effect of the instuction itself, a module $M'$ of end-to-end instruction rewriting can be introduced between input and output to transform the raw one $x$ to a refined version $x'$, 
\begin{equation}
    x' = M'(x)
\end{equation}
which is trained with heuristically synthesized data, conducted by \citet{cheng-etal-2024-black}. Despite improvement, its impact is bottlenecked by data quality.
This paper accordingly design a new pipeline of instruction synthesis, as well as effective training implementation and useful resources.

\section{Methodology}
P-Aligner is a lightweight but powerful module for instruction optimization, targetted on aligning user input to human preference. 
In this part, we first introduce a principled instruction-synthesis pipeline that generates high-quality, preference-oriented rewrite pairs (\S\ref{subsec:synthesis}), and then detail how P-Aligner is trained and deployed (\S\ref{subsec:training}).

\subsection{Principled Instruction Synthesis}
\label{subsec:synthesis}
An instruction can be flawed, e.g., ambiguous, disrespectful, or incomplete, and propagate into poor model outputs.
Effective instruction improvement requires (i) a clear direction toward human preference, and (ii) exhaustive yet efficient exploration of the (near-)paraphrase space, while both sides are absent and contribute to sub-optimal results.

The objective here is to synthesize new instructions compensated along certain principles with given seed instructions. 
Similar to \citet{wang2023promptagent}, we address these gaps with an MCTS-based pipeline guided by explicit \emph{alignment principles} and re-design the inner components, as shown in Figure~\ref{fig:head}.

Concretely, by treating each instruction as a node, a single transition between nodes can be defined. Given an rewriter $O$, a modified instruction $x'$ can be obtained with the raw instruction $x$ and principle $r \in S$: 
\begin{equation}
\label{eq:optimization}
    x' \sim O(x, r)
\end{equation}
Here the set $S$ maintaining all principles define the action space, and the principles are designed as atomic actions, and can be combined through multiple transitions. 

\paragraph{Simulation}
A key design aspect is that every node represents a terminal state, as each transition produces a completed instruction to return.
Consequently, the standard MCTS phase of Simulation is adapted. 
To be specific, if a new node $N$ is to be expanded, initially it should go through a rollout process to a terminal node, and then the reward is calculated to score $N$. However, since $N$ has already a terminal node in our design, this stage is performed in-place, and the corresponding score can be computed directly. Another challenge is how to properly score an instruction, since prevalent reward models only score responses.
Here we define rewarding instruction $x_i$ using its corresponding responses $\textbf{y}_i$ from a local LLM $M$,
\begin{equation}
    \textbf{y}_i \sim M(x_i)
\end{equation}
\begin{equation}
    r_i = \frac{1}{\left| \textbf{y}_i \right|} \sum_{y \in \textbf{y}_i} R(x_0, {y})
\end{equation}
where $R$ is a reward model to score each response ${y}_{i,j}$. 
The instruction reward $r_i$ is the average of multiple response rewards to reduce randomness, as well as based on the original instruction $x_0$, throughout the tree search process, to ensure fairness in comparing different nodes.

\paragraph{Backpropagation}
Next, the reward is backpropagated to update the Q-values of all its ancestors along the full path to the root in the search tree. For a node $N$ containing $p_i$, paired with its child nodes $\{N'_j\}$, we define computing Q-value of $N$ as follows:
\begin{equation}
    Q(N) = \frac{1}{1 + \left|\{N'_j\} \right|} \left (r_i + \sum_{N'_j} Q(N'_j)\right )
\end{equation}
Note that the reward $r_i$ of $p_i$ is also considered, since directly returning $p_i$ can also be an implicit transition, and $Q(N) = r_i$ if $N$ is a leaf node. 

\begin{table}[t]
    \centering
    \resizebox{\linewidth}{!}{
    \begin{tabular}{lcc}
        \toprule
        \textbf{Sources} & \textbf{\# Instructions} & \textbf{Category} \\
        \midrule
        FalseQA~\citep{hu-etal-2023-wont} & 250 & Honesty \\
        TruthfulQA~\citep{lin-etal-2022-truthfulqa} & 250 & Honesty \\
        FLAN~\citep{pmlr-v202-longpre23a} & 300 & Helpfulness \\
        HH-RLHF~\citep{bai2022training} & 1500 & Harmlessness \\
        UltraChat~\citep{ding-etal-2023-enhancing} & 1800 & Helpfulness \\
        ShareGPT~\citep{chiang2023vicuna} & 3000 & Helpfulness \\
        GSM-RFT & 300 & Math \\
        Math50k-camel & 300 & Math \\
        MATH~\citep{hendrycks2021measuring} & 300 & Math \\
        Glaive-code-assistant & 2000 & Coding \& Debugging \\
        \bottomrule
    \end{tabular}}
    \caption{Statistics of the seed instruction sources in UltraPrompt.}
    \label{tab:ultraprompt_end_to_end}
\end{table}

\paragraph{Selection and Expansion}
The Q-values updated above then 
With Q-values as updated above, the search process turns to select the next node to explore, progressively moving toward higher-reward regions of the instruction space.
For the parent node $N$ whose children are $\{N'_j\}$, the child to visit next is chosen via the UCB rule
\begin{equation}
    N'_{\text{next}} = \arg\max_{N'_j} \left ( Q(N'_j) + c \sqrt{\frac{\log V(N)}{V(N'_j)}} \right )
\end{equation}
where $V(N)$ records visit counts to node $N$, and the constant $c$ balances exploration versus exploitation. 
Expansion is then triggered whenever at least one unexplored action remains for $N$, and a new instruction is produced by applying Equation~\ref{eq:optimization} with the selected principle, yielding a fresh node without considering its parent state is terminal. This ensures the search continues rather than terminating instantly at the root node.
\begin{figure}[t]
    \centering
    \includegraphics[width=\linewidth]{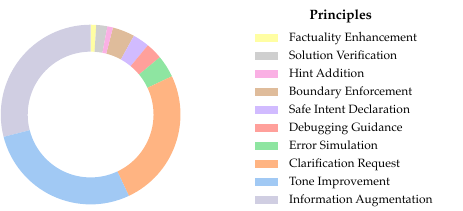} 
    \caption{Distribution of principles in the \textit{single-step} split of UltraPrompt.}
    \label{fig:pancake}
\end{figure}

\begin{table*}[t]
    \centering
    \resizebox{\linewidth}{!}{
    \begin{tabular}{lc ccc ccc ccc ccc}
        \toprule
        \multirow{2}{*}{\textbf{Model}} & \multirow{2}{*}{\textbf{Method}} & \multicolumn{3}{c}{\textbf{Vicuna Eval (VE)}} & \multicolumn{3}{c}{\textbf{Self-instruct Eval (SE)}} & \multicolumn{3}{c}{\textbf{Dolly Eval (DE)}} & \multicolumn{3}{c}{\textbf{BPO Test (BT)}} \\
        \cmidrule(lr){3-5} \cmidrule(lr){6-8} \cmidrule(lr){9-11} \cmidrule(lr){12-14} 
        & & \textbf{Win}$\uparrow$ & \textbf{Tie} & \textbf{Lose}$\downarrow$ & \textbf{Win}$\uparrow$ & \textbf{Tie} & \textbf{Lose}$\downarrow$ & \textbf{Win}$\uparrow$ & \textbf{Tie} & \textbf{Lose}$\downarrow$ & \textbf{Win}$\uparrow$ & \textbf{Tie} & \textbf{Lose}$\downarrow$ \\
        \midrule

        \multirow{3}{*}{GPT-4-turbo} 
         & Normal & 10.00 & 1.25 & 88.75 & 7.94 & 11.11 & 80.95 & 4.50 & 14.50 & 81.00 & 7.50 & 14.00 & 78.50 \\ 
         & BPO & 21.25 & 2.50 & 76.25 & 19.05 & 8.73 & 72.22 & 27.00 & 13.00 & 60.00 & 22.50 & 15.00 & 62.50 \\ 
         & \cellcolor[RGB]{235,235,235} P-Aligner & \cellcolor[RGB]{235,235,235}\textbf{50.00} & \cellcolor[RGB]{235,235,235}\textbf{3.75} & \cellcolor[RGB]{235,235,235}\textbf{46.25} & \cellcolor[RGB]{235,235,235}\textbf{54.37} & \cellcolor[RGB]{235,235,235}\textbf{3.97} & \cellcolor[RGB]{235,235,235}\textbf{41.66} & \cellcolor[RGB]{235,235,235}\textbf{68.50} & \cellcolor[RGB]{235,235,235}\textbf{6.50} & \cellcolor[RGB]{235,235,235}\textbf{25.00} & \cellcolor[RGB]{235,235,235}\textbf{57.50} & \cellcolor[RGB]{235,235,235}\textbf{12.50} & \cellcolor[RGB]{235,235,235}\textbf{30.00}  \\ \midrule

        \multirow{3}{*}{Gemma-2-SimPO} 
         & Normal & 75.00 & 2.50 & 22.50 & 59.52 & 7.54 & 32.94 & 54.50 & 9.00 & 36.50 & 59.00 & 14.00 & 27.00 \\ 
         & BPO & 78.75 & 2.50 & 18.75 & 65.87 & 6.35 & 27.78 & 63.00 & 9.00 & 28.00 & 62.00 & 11.50 & 26.50 \\ 
         & \cellcolor[RGB]{235,235,235} P-Aligner & \cellcolor[RGB]{235,235,235}\textbf{83.75} & \cellcolor[RGB]{235,235,235}\textbf{0.00} & \cellcolor[RGB]{235,235,235}\textbf{16.25} & \cellcolor[RGB]{235,235,235}\textbf{72.22} & \cellcolor[RGB]{235,235,235}\textbf{4.76} & \cellcolor[RGB]{235,235,235}\textbf{23.02} & \cellcolor[RGB]{235,235,235}\textbf{78.00} & \cellcolor[RGB]{235,235,235}\textbf{4.00} & \cellcolor[RGB]{235,235,235}\textbf{18.00} & \cellcolor[RGB]{235,235,235}\textbf{75.50} & \cellcolor[RGB]{235,235,235}\textbf{9.50} & \cellcolor[RGB]{235,235,235}\textbf{15.00} \\ \midrule

        \multirow{3}{*}{Llama-3.1-8B}
        & Normal & 0.00 & 0.00 & 100.00 & 1.98 & 0.00 & 98.02 & 1.50 & 3.00 & 95.50 & 1.50 & 1.00 & 97.50 \\
        & BPO & 3.75 & 1.25 & 95.00 & 5.16 & 0.40 & 94.44 & 6.50 & 1.50 & 92.00 & 4.00 & 1.50 & 94.50 \\
        & \cellcolor[RGB]{235,235,235}P-Aligner & \cellcolor[RGB]{235,235,235}\textbf{6.25} & \cellcolor[RGB]{235,235,235}\textbf{1.25} & \cellcolor[RGB]{235,235,235}\textbf{92.50} & \cellcolor[RGB]{235,235,235}\textbf{22.22} & \cellcolor[RGB]{235,235,235}\textbf{1.59} & \cellcolor[RGB]{235,235,235}\textbf{76.19} & \cellcolor[RGB]{235,235,235}\textbf{37.00} & \cellcolor[RGB]{235,235,235}\textbf{3.00} & \cellcolor[RGB]{235,235,235}\textbf{60.00} & \cellcolor[RGB]{235,235,235}\textbf{27.50} & \cellcolor[RGB]{235,235,235}\textbf{2.50} & \cellcolor[RGB]{235,235,235}\textbf{70.00} \\
        \midrule
        \multirow{3}{*}{\quad w/ Best-of-N}
        & Normal & 0.00 & 0.00 & 100.00 & 1.19 & 0.40 & 98.41 & 2.50 & 0.50 & 97.00 & 4.50 & 1.00 & 94.50 \\
        & BPO & 6.25 & 0.00 & 93.75 & 2.78 & 0.79 & 96.43 & 8.50 & 1.00 & 90.50 & 8.00 & 0.50 & 91.50 \\
        & \cellcolor[RGB]{235,235,235}P-Aligner & \cellcolor[RGB]{235,235,235}\textbf{13.75} & \cellcolor[RGB]{235,235,235}\textbf{0.00} & \cellcolor[RGB]{235,235,235}\textbf{86.25} & \cellcolor[RGB]{235,235,235}\textbf{28.97} & \cellcolor[RGB]{235,235,235}\textbf{3.97} & \cellcolor[RGB]{235,235,235}\textbf{67.06} & \cellcolor[RGB]{235,235,235}\textbf{42.50} & \cellcolor[RGB]{235,235,235}\textbf{4.50} & \cellcolor[RGB]{235,235,235}\textbf{53.00} & \cellcolor[RGB]{235,235,235}\textbf{41.00} & \cellcolor[RGB]{235,235,235}\textbf{2.50} & \cellcolor[RGB]{235,235,235}\textbf{56.50} \\
        \midrule
        \multirow{3}{*}{\quad w/ URIAL}
        & Normal & 5.00 & 0.00 & 95.00 & 14.68 & 1.98 & 83.34 & 23.00 & 2.00 & 75.00 & 15.00 & 3.50 & 81.50 \\
        & BPO & 5.00 & 0.00 & 95.00 & 14.68 & 0 & 85.32 & 25.50 & 0.00 & 74.50 & 15.50 & 3.00 & 81.50 \\
        & \cellcolor[RGB]{235,235,235}P-Aligner & \cellcolor[RGB]{235,235,235}\textbf{5.00} & \cellcolor[RGB]{235,235,235}\textbf{0.00} & \cellcolor[RGB]{235,235,235}\textbf{95.00} & \cellcolor[RGB]{235,235,235}\textbf{20.24} & \cellcolor[RGB]{235,235,235}\textbf{1.98} & \cellcolor[RGB]{235,235,235}\textbf{77.78} & \cellcolor[RGB]{235,235,235}\textbf{40.00} & \cellcolor[RGB]{235,235,235}\textbf{1.50} & \cellcolor[RGB]{235,235,235}\textbf{58.50} & \cellcolor[RGB]{235,235,235}\textbf{32.50} & \cellcolor[RGB]{235,235,235}\textbf{3.00} & \cellcolor[RGB]{235,235,235}\textbf{64.50} \\
        \bottomrule
    \end{tabular}
    }
    \caption{Results across Vicuna Evaluation, Self-instruct Evaluation, Dolly Evaluation and BPO Test. Higher win rates represents better performance.}
    \label{tab:main_results}
\end{table*}
\begin{figure*}[t]
    \centering
    \includegraphics[width=1\linewidth]{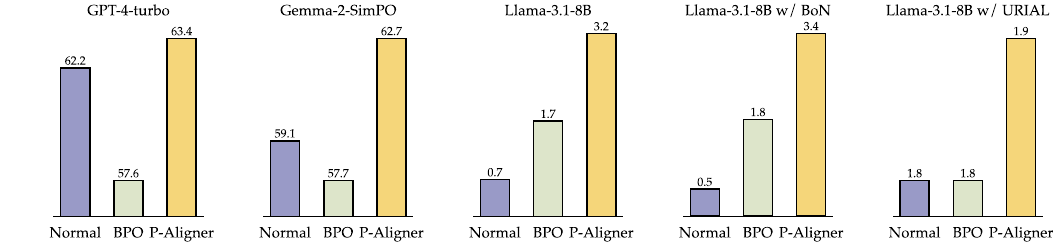} 
    \caption{Results on ArenaHard. Higher scores represents better performance.}
    \label{fig:arena}
\end{figure*}

\subsection{P-Aligner}
\label{subsec:training}
With the above pipeline, each seed instruction can be rewritten into a set of derived ones. In this part, we elaborate on how we leverage it to acquire P-Aligner with preference learning. 

We first propose \textbf{UltraPrompt}, a preference dataset synthesized through the above pipeline. 
It contains 10000 seed instructions selected from various sources: UltraFeedback~\citep{cui2024ultrafeedback}, HH-RLHF~\citep{bai2022training}, Glaive-code-assistant\footnote{https://huggingface.co/datasets/glaiveai/glaive-code-assistant}, and MathInstruct~\citep{yue2024mammoth}, which is completed according to N-grams diversity in \citet{song-etal-2024-scaling} to cover various domains, as shown in Table~\ref{tab:ultraprompt_end_to_end}.
We further exploit the scores of instrctions during data synthesis to filter out the best/worst version of derived instructions as the chosen/rejected targets, thus assembling a contrastive sample with the seed instruction: \texttt{(seed, chosen, rejected)} for preference learning. \textbf{P-Aligner} is finally gained with DPO~\cite{rafailov2024direct}, an efficient algorithm to rapidly align LLMs to customized preference. For implementation, its end-to-end production of principled instructions will reduce the cost in multiple aspects, as discussed in \S\ref{sec:consumption}.

Furthermore, we acquire \textbf{SinglePO}, a local-implemented alternative rewriter $O$ used in our proposed data synthesis pipeline to reduce the cost in financial and time overhead. It derives from an additional \textit{single-step} split of UltraPrompt, where we reuse the 10000 search trees and collect all 104602 positive transitions, i.e., where the next instruction has a higher reward than the current instruction. The distribution of principles in the \textit{single-step} split is shown in Figure~\ref{fig:pancake}. 

\begin{table*}[t]
    \centering
    \resizebox{\linewidth}{!}{
    \begin{tabular}{lc ccc ccc ccc ccc}
        \toprule
        \multirow{2}{*}{\textbf{Model}} & \multirow{2}{*}{\makecell{\textbf{Range of}\\\textbf{Sampling}}} & \multicolumn{3}{c}{\textbf{Vicuna Eval (VE)}} & \multicolumn{3}{c}{\textbf{Self-instruct Eval (SE)}} & \multicolumn{3}{c}{\textbf{Dolly Eval (DE)}} & \multicolumn{3}{c}{\textbf{BPO Test (BT)}} \\
        \cmidrule(lr){3-5} \cmidrule(lr){6-8} \cmidrule(lr){9-11} \cmidrule(lr){12-14}
        & & \textbf{Win}$\uparrow$ & \textbf{Tie} & \textbf{Lose}$\downarrow$ & \textbf{Win}$\uparrow$ & \textbf{Tie} & \textbf{Lose}$\downarrow$ & \textbf{Win}$\uparrow$ & \textbf{Tie} & \textbf{Lose}$\downarrow$ & \textbf{Win}$\uparrow$ & \textbf{Tie} & \textbf{Lose}$\downarrow$ \\
        \midrule
        \multirow{4}{*}{GPT-4-turbo}
        & 2 steps & 16.25 & 7.50 & 76.25 & 26.59 & 13.49 & 59.92 & 26.50 & 16.50 & 57.00 & 23.00 & 17.00 & 60.00 \\
        & 11 steps & 48.75 & 10.00 & 41.25 & 45.24 & 13.10 & 41.66 & 63.50 & 9.00 & 27.50 & 52.50 & 17.50 & 30.00 \\
        & \cellcolor[RGB]{235,235,235}max steps & \cellcolor[RGB]{235,235,235}\textbf{50.00} & \cellcolor[RGB]{235,235,235}\textbf{3.75} & \cellcolor[RGB]{235,235,235}\textbf{46.25} & \cellcolor[RGB]{235,235,235}\textbf{54.37} & \cellcolor[RGB]{235,235,235}\textbf{3.97} & \cellcolor[RGB]{235,235,235}\textbf{41.66} & \cellcolor[RGB]{235,235,235}\textbf{68.50} & \cellcolor[RGB]{235,235,235}\textbf{6.50} & \cellcolor[RGB]{235,235,235}\textbf{25.00} & \cellcolor[RGB]{235,235,235}\textbf{57.50} & \cellcolor[RGB]{235,235,235}\textbf{12.50} & \cellcolor[RGB]{235,235,235}\textbf{30.00} \\
        & random & 41.25 & 3.75 & 55.00 & 46.83 & 6.75 & 46.42 & 58.50 & 10.00 & 31.50 & 48.50 & 10.00 & 41.50 \\
        \midrule
        \multirow{4}{*}{Gemma-2-SimPO}
        & 2 steps & 75.00 & 0 & 25.00 & 63.89 & 7.54 & 28.57 & 65.00 & 7.50 & 27.50 & 63.50 & 11.50 & 25.00 \\
        & 11 steps & 76.25 & 12.50 & 11.25 & 66.67 & 7.54 & 25.79 & 73.50 & 6.50 & 19.50 & 71.00 & 8.50 & 20.50 \\
        & \cellcolor[RGB]{235,235,235}max steps & \cellcolor[RGB]{235,235,235}\textbf{83.75} & \cellcolor[RGB]{235,235,235}\textbf{0.00} & \cellcolor[RGB]{235,235,235}\textbf{16.25} & \cellcolor[RGB]{235,235,235}\textbf{72.22} & \cellcolor[RGB]{235,235,235}\textbf{4.76} & \cellcolor[RGB]{235,235,235}\textbf{23.02} & \cellcolor[RGB]{235,235,235}\textbf{78.00} & \cellcolor[RGB]{235,235,235}\textbf{4.00} & \cellcolor[RGB]{235,235,235}\textbf{18.00} & \cellcolor[RGB]{235,235,235}\textbf{75.50} & \cellcolor[RGB]{235,235,235}\textbf{9.50} & \cellcolor[RGB]{235,235,235}\textbf{15.00} \\
        & random & 78.75 & 1.25 & 20.00 & 69.44 & 3.97 & 26.59 & 76.50 & 4.00 & 19.50 & 65.00 & 8.00 & 27.00 \\
        \bottomrule
    \end{tabular}
    }
    \caption{Results of P-Aligner with different ways of sampling preferred instructions, including random sampling and sampling with highest rewards from various limited steps in search trees. The maximum steps is 20.}
    \label{tab:data_quality}
\end{table*}

\section{Experiments}
\subsection{Evaluation Setup}
Following \citet{cheng-etal-2024-black}, we take different benchmarks in preference learning and instruction following to measure the performance of each setting, including Vicuna Evaluation~\citep{chiang2023vicuna}, Self-instruct Evaluation~\citep{wang-etal-2023-self-instruct}, Dolly Evaluation~\citep{conover2023free}, BPO Test. We further introduce ArenaHard~\citep{li2024crowdsourced}, a popular benchmark to accurately judge the model capabilities of instruction following.

For ArenaHard, we directly take its initial output score as the metric, while for the other 4 benchmarks, the win/tie/lose-rates, i.e., rates of generated responses better/comparable/worse against baseline ones, computed in double directions, are the common metrics, which has been widely used in prior works~\citep{wang2023large}. We also set GPT-4o responses as the baseline for calculating the win-rate to facilitate reproducibility.

\subsection{Baselines} 
We compare P-Aligner with two baselines: using original instructions~(\textbf{Normal}) or instructions optimized by \textbf{BPO}~\citep{cheng-etal-2024-black}. Such comparisons are conducted on various models and methods, including the close-source GPT-4-turbo, the open-source instruct model Gemma-2-SimPO~\citep{meng2024simpo}, as well as tuning-free methods such as Best-of-N and URIAL~\citep{lin2024the} based on Llama-3.1-8B~\citep{dubey2024llama}. All experiments have excluded system prompts to eliminate the effect of external factors.

\subsection{Implementation Details}
In the process of instruction synthesis, we deploy GPT-4 as the default rewriter $O$. Each principle is mapped into a description (see Appendix~A) and embedded in the input to $O$.
We take Llama-3.1-8B as the response generator $M$ for given instructions. Once a simulation stage begins, there are 3 responses sampled from $M$, and then the open-source reward model, ArmoRM-Llama3-8B-v0.1~\citep{wang2024interpretable}, is leveraged to score these responses, which indirectly reflects the quality of the given instructions. 
The exploration weight to balance exploration and exploitation in MCTS is set to 0.1 to fit the scale of outputs from the reward model.

We train P-Aligner and SinglePO from Llama-3.2-3B-Instruct~\citep{grattafiori2024llama}, which are completed by LlamaFactory~\citep{zheng-etal-2024-llamafactory}. Inference with the two models is conducted with greedy search to ensure reproducibility. 

\begin{figure}[t]
    \centering
    \includegraphics[width=\linewidth]{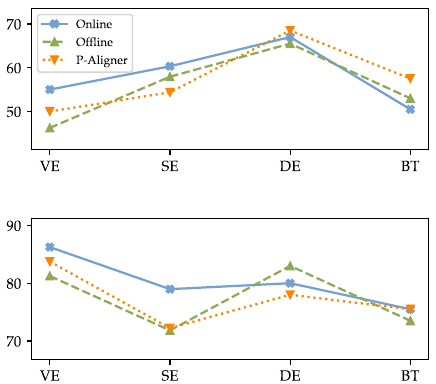}
    \caption{Comparisons among different search strategies. Upper: GPT-4-turbo. Lower: Gemma-2-SimPO. VE, SE, DE and BT denote Vicuna Evaluation, Self-instruct Evaluation, Dolly Evaluation and BPO Test, respectively.}
    \label{fig:search_strategy}
\end{figure}

\subsection{Main Results}
Table~\ref{tab:main_results} summarizes the benchmark results, where P-Aligner consistently surpasses both the vanilla (Normal) and BPO baselines across all tested settings.
On GPT-4-turbo, P-Aligner raises the win rate by 28.35\% overall, with especially large gains of 28.75\% on Vicuna Eval and 35.32\% on Self-Instruct Eval (SE).
Gemma-2-SimPO likewise benefits, showing an 8.69\% average improvement.
Even on the more challenging ArenaHard benchmark, as shown in Figure~\ref{fig:arena}, P-Aligner delivers further score increases across different models and methods.
These results confirm its robust capacity to enhance LLM preference alignment.

We also find that P-Aligner and BPO both bring fewer improvements on ArenaHard~(AH) compared to other benchmarks. We investigate the instruction in AH which can be more specific and clear compared to other benchmarks, which may limit the room for further improvement. 
Moreover, the effectiveness of instruction optimization also depends on context. For example, since URIAL maintains a static context to shift the distribution of LLMs, the improvements from P-Aligner and BPO on it are partially offset.

\subsection{Ablation Study}
\subsubsection{Effect of Training Data}
A crucial aspect of P-Aligner is collecting the training data for UltraPrompt. The default way selects the nodes with the highest and lowest rewards within each completed search tree. 
To evaluate the influence of data quality, we test several alternative collection methods, including random sampling (selecting preferred samples from random nodes) and sampling from truncated search trees at 2 and 11 steps.

Table~\ref{tab:data_quality} shows the results that with larger trees, P-Aligner performs better, indicating the robustness of our proposed instruction synthesis pipeline and the effectiveness of the reward mechanism in producing better instructions.
On the other hand, the effect of random sampling is not stable and sometimes the lowest, for which we infer that fluctuations in data quality hinder effective learning.

\subsubsection{Comparisons among Different Search Strategies}
We evaluate the performance of P-Aligner alongside two alternative search strategies: \textbf{online} search, which utilizes the proposed pipeline with GPT-4 for instruction synthesis, and \textbf{offline} search, which employs SinglePO for the same purpose. Experiments are conducted on both GPT-4-turbo and Gemma-2-SimPO and involve the four benchmarks from \cite{cheng-etal-2024-black}.

As exhibited in Figure~\ref{fig:search_strategy}, the performance of these methods are closely matched, with their relative rankings varying across different benchmarks. For instance, on the Vicuna Eval~(VE), the online method leads with a 5 percentage point advantage over P-Aligner with GPT-4-turbo, while outperforming the offline approach by 8.75 points. However, this pattern reverses on the Dolly Eval~(DE), where P-Aligner claims the top spot with a 68.50 point, over both the online and offline search.
Similar fluctuations also exist with Gemma-2-SimPO, where P-Aligner lead the group for BPO Test~(BT) while the online search stays at the top position for Self-instruct Eval~(SE). 
Such close performance underscore the robustness and versatility of P-Aligner, and after considering the cost of implementation, P-Aligner become the most recommended choice, which we discuss in \S\ref{sec:consumption}.

\section{Discussion}
\subsection{Effect of Iterative Optimization}
An interesting observation in \citet{cheng-etal-2024-black} is that repeatedly applying BPO can yield further gains.
While empirically useful, this practice is economically unattractive: each additional round incurs additional latency of the rewriter.

We hypothesize that the diminishing-returns improvement of iteratively running BPO stems from the corpus in its training. To be specific, each refined instruction is produced by a single heuristic rewrite step, whose direction is therefore implicit and slight.
Consequently, multi-time BPO also functions like a low-resolution search process. In contrast, resources in UltraPrompt, which support P-Aligner, are already near-optimal with iteratively search in data synthesis while requiring no human annotation. We check this point by replicate the iterative experiment with P-Aligner, using Gemma-2-SimPO and measuring performance via win-rate on four benchmarks involved in \citet{cheng-etal-2024-black}.

\begin{figure}[t]
    \centering
    \includegraphics[width=1\linewidth]{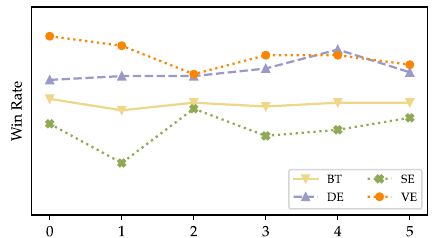} 
    \caption{Effect of iterative instruction optimization with P-Aligner, where P-Aligner can almost achieve the highest benefit without multiple runs. The X-axis represents the index of iteration.}
    \label{fig:iter}
\end{figure}

\begin{table}[t]
\centering
\resizebox{\linewidth}{!}{
    \begin{tabular}{lcccc}
    \toprule
    \textbf{Strategies} & \textbf{Free Use} & \textbf{Local Use} & \textbf{E2E} & \textbf{Time Overhead} \\ \midrule
    Online & \xmark & \xmark & \xmark & 5300 ms \\
    Offline & \cmark & \cmark & \xmark & 3920 ms \\
    \rowcolor[RGB]{235,235,235}P-Aligner & \cmark & \cmark & \cmark & \textbf{108 ms} \\
    \bottomrule
    \end{tabular}
}
\caption{Comparisons among different search strategies in multiple aspects of implementation.}
\label{tab:comparison}
\end{table}

\begin{figure}[t]
    \centering
    \includegraphics[width=\linewidth]{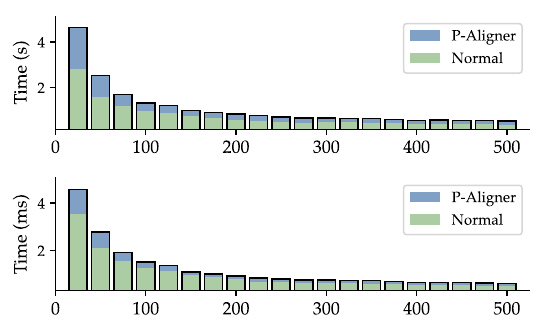}
    \caption{Comparisons of time between normal inference and inference with P-Aligner. The X-axis represents the number of batch-submitted queries. Upper: time consumed per query. Lower: time consumed per token.}
    \label{fig:time_overhead}
\end{figure}

Figure~\ref{fig:iter} illustrates the results.
Unlike BPO that shows gradual improvement with additional passes, P-Aligner exhibits no consistent trend across iterations. For example, performance on BPO Test and Dolly Eval remains stable, while scores on Self-Instruct Eval and Vicuna Eval show clear fluctuation, or even decrese~(in Vicuna Eval).
These indicates that P-Aligner delivers near-optimal instructions in a single step, eliminating the need and consumption from iterative refinement.

\subsection{Analysis of Consumption}
\label{sec:consumption}
Compared with direct inference on raw user query, introducing instruction optimization in advance inevitably brings additional overhead in multiple aspects, such as time overhead and financial cost.
To quantify these trade-offs, we compare different search strategies discussed before with P-Aligner, along four axes: financial cost, local use, end-to-end execution and time overhead. Results are included in Table~\ref{tab:comparison}, where GPT-4-based online search is the most expensive and slowest due to frequent API calls. 
In contrast, offline search with our proposed SinglePO can achieve comparable alignment quality while lowering cost, security (with local implementation) and time consumption, making it a valuable alternative.
P-Aligner, however, emerges as the most economical: it is locally deployable, off-the-shelf, and executes in an end-to-end manner, eliminating both API fees and multi-stage latency.

To further assess the marginal time cost introduced by P-Aligner, we measure (i)~average response time per query and (ii)~average decoding time per token on Gemma-2-SimPO over 25–500 instructions sampled from ArenaHard.
Figure~\ref{fig:time_overhead} shows that the relative overhead is most significant for small batch sizes, and as the number of queries grows, the amortized cost rapidly diminishes.
Moreover, since P-Aligner is lightweight, the relative overhead is expected to shrink further when paired with larger models, whose base inference time dominates the total budget.

\section{Conclusion}
LLMs often fail to align with human preference because the instructions themselves are ambiguous, biased, or poorly phrased, raising the urgent need of a low-overhead but effective pre-alignment mechanism on instructions.
In this work, we present a novel pipeline that couples Monte-Carlo Tree Search with preference-aligned principles to synthesize human-preferred instructions.
From this pipeline we derive UltraPrompt, a high-quality preference dataset of synthesized instruction, enabling to effectively train P-Aligner, a lightweight end-to-end module that refines raw instructions in a single forward pass.
With UltraPrompt, we also introduce SinglePO, a single-step variant that allows the proposed data synthesis pipeline to be execute entirely on local hardware without sharp loss of performance.
Extensive evaluations across different benchmarks, baselines and models confirm consistent gains from P-Aligner in preference learning while incurring minimal additional overhead.
Together, these contributions offers a promising direction to establish instruction-level pre-alignment as a practical and scalable complement in preference learning.

\section*{Ethical Statement}
The data used in evaluation and UltraPrompt may include sensitive information, such as misleading content or offensive instructions. Such data does not represent our attitudes, and should be handled carefully to avoid potential harm. We request any potential users treat it responsibly without any use or distribution outside of research contexts.

\appendix
\section{Principles}
\label{sec:appendix_principles}
The principles are pre-defined to cover commonly recognized positive aspects for achieving human preference. It involves multiple domains:
Harmlessness, Helpfulness, Honesty, Coding \& Debugging, and Math. Each domain contains several priciples.
Importantly, the principles of Helpfulness are intended to be universally applicable across all defined domains.
A detailed categorization is presented below:\\
\\
\textbf{Harmlessness}:
\begin{itemize}
    \item \textbf{Safe Intent Declaration} - adding a safety-oriented preface (e.g., "Please respond respectfully and avoid harmful or unethical content...") requiring the response to meet ethical guidelines.
    \item \textbf{Boundary Enforcement} - appending explicit refusal instructions for any content violating privacy/ethics (e.g., "If this request involves offensive content, decline politely").
\end{itemize}
\textbf{Helpfulness}:
\begin{itemize}
    \item \textbf{Clarification Request} - making it more clear and instructive, leading the agent to answer it in detail.
    \item \textbf{Information Augmentation} - making it more detailed and informative, such as adding more background information and so on, which may help the agent better understand the content.
    \item \textbf{Tone Improvement} - improving its tone to be more polite, helpful, honest, and friendly.
\end{itemize}
\textbf{Honesty}:
\begin{itemize}
    \item \textbf{Factuality Enhancement} - encouraging objective facts instead of fake material and subjective interpretations.
\end{itemize}
\textbf{Coding \& Debugging}:
\begin{itemize}
    \item \textbf{Error Simulation} - adding requests about considering common bugs or edge cases related to the prompt.
    \item \textbf{Debugging Guidance} - offering step-by-step debugging instructions, error analysis, and troubleshooting tips to resolve coding issues.
\end{itemize}
\textbf{Math}:
\begin{itemize}
    \item \textbf{Hint Addition} - introducing hints, examples or multi-step pre-thinking to reduce the difficulty of problems.
    \item \textbf{Solution Verification} - asking the agent to verify the correctness of their response or provide a detailed explanation of their reasoning.
\end{itemize}

\section{Further Explanation of Evaluation Settings}
In this work, we evaluate our methods on five instruction-following benchmarks.
Four of them are also adopted by \citet{cheng-etal-2024-black}, which we list as follows:
\begin{itemize}
\item \textbf{Dolly Eval} contains 200 samples drawn from the Dolly dataset~\citep{conover2023free} to test general instruction following performance.
\item \textbf{Vicuna Eval}~\citep{chiang2023vicuna} contains 80 samples spanning role-play, common-sense, creative writing, coding, mathematics, and other categories.
\item \textbf{Self-Instruct Eval}~\citep{wang-etal-2023-self-instruct} contains 252 carefully crafted samples, each pairing a context with an instruction.
\item \textbf{BPO Test} contains 200 samples introduced by \citet{cheng-etal-2024-black}.
\end{itemize}
We further include \textbf{ArenaHard}~\citep{li2024crowdsourced}, a popular benchmark containing 500 challenging prompts and an automated evaluation framework that yields a scalar score of instruction-following performance.

Beyond vanilla inference, we further equip Llama-3.1-8B with two decoding-time alignment strategies.  
\textbf{Best-of-N} samples \(N\) candidate responses and returns the highest-scoring one as judged by ArmoRM-Llama3-8B-v0.1.  
\textbf{URIAL}~\citep{lin2024the} steers the decoding distribution via in-context learning on human-preferred demonstrations.  
Both methods function at generation time and are orthogonal to BPO and P-Aligner, which work by refining the prompts, so they can be well aggregated for a more comprehensive evaluation and robust validation of our findings.

\section{Prompt Templates}
Please see Figure~\ref{fig:prompt_1}, \ref{fig:prompt_2} and \ref{fig:prompt_3} for reference.

\begin{figure*}[h]
  \centering
  \resizebox{1\textwidth}{!}{
  \begin{tcolorbox}
    The user query to be paraphrased is \{The given query\}. 
    
    You should optimize this query by \{Content of the selected rule\}. 
    
    You should also return the optimized version directly, without any prefix.
  \end{tcolorbox}
  }
  \caption{
  The prompt template used for GPT-4 to improve the given query.}
  \label{fig:prompt_1}
\end{figure*}

\begin{figure*}[h]
  \centering
  \resizebox{1\textwidth}{!}{
  \begin{tcolorbox}
    You are an expert prompt engineer. Please help me optimize this prompt to get better response:\\

    [The Start of Raw Prompt]

    \{The given query\}

    [The End of Raw Prompt]\\

    You should optimize this prompt by \{Content of the selected rule\}.
  \end{tcolorbox}
  }
  \caption{
  The prompt template used for SinglePO to improve the given query.}
  \label{fig:prompt_2}
\end{figure*}

\begin{figure*}[h]
  \centering
  \resizebox{1\textwidth}{!}{
  \begin{tcolorbox}
    You are an expert prompt engineer. Please help me optimize this prompt to get better response:\\

    [The Start of Raw Prompt]

    \{The given query\}

    [The End of Raw Prompt]
  \end{tcolorbox}
  }
  \caption{
  The prompt template used for MCTS-p to improve the given query.}
  \label{fig:prompt_3}
\end{figure*}

\section{Cases}
Please see Table~\ref{tab:case_1} and \ref{tab:case_2} for reference.

\begin{table*}[h]
\centering
    \begin{tabularx}{\linewidth}{>{\hsize=0.2\hsize}X>{\hsize=0.8\hsize}X}
    \toprule
    \textbf{Raw Prompt} & \textbf{User:} How can governments utilize fiscal and monetary policies to combat economic recessions? \newline\newline \textbf{Llama-3.1-8B:} Governments can utilize fiscal and monetary policies to combat economic recessions by increasing government spending, reducing taxes, and increasing the money supply. These policies can help stimulate the economy and create jobs, which can help to reduce the severity of a recession.\\
    \midrule
    \textbf{Prompt by BPO} & \textbf{User:} How can governments utilize fiscal and monetary policies to combat economic recessions? Provide a detailed explanation of the specific measures governments can implement through fiscal and monetary policies to address economic downturns. Include examples of how these policies can impact the economy, such as increasing demand and reducing costs. \newline\newline \textbf{Llama-3.1-8B:} Governments can utilize fiscal and monetary policies to combat economic recessions by implementing measures such as increasing government spending, reducing taxes, and providing stimulus packages. These policies can help increase demand in the economy and reduce costs, which can help stimulate economic growth and reduce unemployment. For example, increasing government spending on infrastructure projects can create jobs and boost economic activity, while reducing taxes can put more money in the hands of consumers, who can then spend it on goods and services, further stimulating the economy. Additionally, providing stimulus packages can provide financial support to businesses and individuals, helping to keep them afloat during difficult times. Overall, these policies can help governments combat economic recessions and promote economic stability and growth.\\
    \bottomrule
    \end{tabularx}
\end{table*}

\begin{table*}[h]
\centering
    \begin{tabularx}{\linewidth}{>{\hsize=0.2\hsize}X>{\hsize=0.8\hsize}X}
    \toprule
    \textbf{Prompt by P-Aligner} & \textbf{User:} Could you kindly provide a detailed explanation on the various fiscal and monetary policies that governments can implement to mitigate the effects of economic downturns or recessions? I'm particularly interested in understanding the specific actions they can take to stimulate economic growth, manage inflation, and maintain financial stability. Your expertise on this topic is greatly appreciated. \newline\newline \textbf{Llama-3.1-8B:} Certainly! Here are some of the key fiscal and monetary policies that governments can use to address economic downturns and recessions:\newline\newline 1. \textbf{Fiscal Stimulus:} Governments can use fiscal stimulus measures such as tax cuts, increased government spending, or direct cash transfers to individuals or businesses to boost consumer spending and investment. This can help stimulate economic growth and create jobs.\newline\newline 2. \textbf{Monetary Policy:} Central banks can use monetary policy tools such as interest rate adjustments, quantitative easing, or open market operations to influence the supply of money and credit in the economy. Lowering interest rates can encourage borrowing and spending, while quantitative easing can increase the money supply and stimulate economic activity.\newline\newline 3. \textbf{Exchange Rate Policy:} Governments can adjust their exchange rates to make their exports more competitive or to attract foreign investment. This can help boost exports and stimulate economic growth.\newline\newline 4. \textbf{Regulatory Policy:} Governments can use regulatory policies to encourage investment and innovation, such as tax incentives for research and development or streamlined permitting processes for new businesses.\newline\newline 5. \textbf{Social Safety Net:} Governments can strengthen their social safety net programs, such as unemployment insurance, food stamps, or housing assistance, to support individuals and families during economic downturns.\newline\newline 6. \textbf{Infrastructure Investment:} Governments can invest in infrastructure projects such as roads, bridges, and public transportation to create jobs and stimulate economic growth.\newline\newline 7. \textbf{Trade Policy:} Governments can use trade policies such as tariffs or import quotas to protect domestic industries or to encourage exports.\newline\newline 8. \textbf{Financial Sector Regulation:} Governments can regulate the financial sector to prevent excessive risk-taking and to ensure that financial institutions are well-capitalized and able to withstand economic shocks.\newline\newline 9. \textbf{Public Debt Management:} Governments can manage their public debt levels to ensure that they are sustainable and do not become a burden on future generations.\newline\newline 10. \textbf{Economic Forecasting:} Governments can use economic forecasting models to anticipate economic downturns and recessions and to develop appropriate policy responses.\newline\newline Overall, governments have a range of fiscal and monetary policy tools at their disposal to address economic downturns and recessions. The specific policies that are most effective will depend on the nature and severity of the economic downturn, as well as the specific circumstances of each country or region.\\
    \bottomrule
    \end{tabularx}
    \caption{Case demonstration from Vicuna Eval.}
    \label{tab:case_1}
\end{table*}

\begin{table*}[h]
\centering
    \begin{tabularx}{\linewidth}{>{\hsize=0.2\hsize}X>{\hsize=0.8\hsize}X}
    \toprule
    \textbf{Raw Prompt} & \textbf{User:} Who was the lead singer for the band Aerosmith? \newline\newline \textbf{Gemma-2-SimPO:} The lead singer for the band Aerosmith is \textbf{Steven Tyler}.\\
    \midrule
    \textbf{Prompt by BPO}\newline (identical to \newline the Raw Prompt) & \textbf{User:} Who was the lead singer for the band Aerosmith? \newline\newline \textbf{Gemma-2-SimPO:} The lead singer for the band Aerosmith is \textbf{Steven Tyler}.\\
    \midrule
    \textbf{Prompt by P-Aligner} & \textbf{User:} Could you please provide information on the lead vocalist of the renowned American rock band Aerosmith? \newline\newline \textbf{Gemma-2-SimPO:} The lead vocalist of the renowned American rock band Aerosmith is \textbf{Steven Tyler}. \newline\newline Known for his powerful, raspy vocals, flamboyant stage presence, and signature wide vocal range, Tyler is a defining element of Aerosmith's iconic sound and image. He's not just a vocalist; he's a charismatic frontman who heavily contributes to the band's energetic performances and enduring appeal.\\
    \bottomrule
    \end{tabularx}
    \caption{Case demonstration from Dolly Eval.}
    \label{tab:case_2}
\end{table*}


\begin{thebibliography}{38}
\providecommand{\natexlab}[1]{#1}

\bibitem[{Achiam et~al.(2023)Achiam, Adler, Agarwal, Ahmad, Akkaya, Aleman, Almeida, Altenschmidt, Altman, Anadkat et~al.}]{achiam2023gpt}
Achiam, J.; Adler, S.; Agarwal, S.; Ahmad, L.; Akkaya, I.; Aleman, F.~L.; Almeida, D.; Altenschmidt, J.; Altman, S.; Anadkat, S.; et~al. 2023.
\newblock Gpt-4 technical report.
\newblock \emph{arXiv preprint arXiv:2303.08774}.

\bibitem[{Bai et~al.(2022)Bai, Jones, Ndousse, Askell, Chen, DasSarma, Drain, Fort, Ganguli, Henighan et~al.}]{bai2022training}
Bai, Y.; Jones, A.; Ndousse, K.; Askell, A.; Chen, A.; DasSarma, N.; Drain, D.; Fort, S.; Ganguli, D.; Henighan, T.; et~al. 2022.
\newblock Training a helpful and harmless assistant with reinforcement learning from human feedback.
\newblock \emph{arXiv preprint arXiv:2204.05862}.

\bibitem[{Brown et~al.(2020)Brown, Mann, Ryder, Subbiah, Kaplan, Dhariwal, Neelakantan, Shyam, Sastry, Askell et~al.}]{brown2020language}
Brown, T.; Mann, B.; Ryder, N.; Subbiah, M.; Kaplan, J.~D.; Dhariwal, P.; Neelakantan, A.; Shyam, P.; Sastry, G.; Askell, A.; et~al. 2020.
\newblock Language models are few-shot learners.
\newblock \emph{Advances in neural information processing systems}, 33: 1877--1901.

\bibitem[{Cheng et~al.(2024)Cheng, Liu, Zheng, Ke, Wang, Dong, Tang, and Huang}]{cheng-etal-2024-black}
Cheng, J.; Liu, X.; Zheng, K.; Ke, P.; Wang, H.; Dong, Y.; Tang, J.; and Huang, M. 2024.
\newblock Black-Box Prompt Optimization: Aligning Large Language Models without Model Training.
\newblock In Ku, L.-W.; Martins, A.; and Srikumar, V., eds., \emph{Proceedings of the 62nd Annual Meeting of the Association for Computational Linguistics (Volume 1: Long Papers)}, 3201--3219. Bangkok, Thailand: Association for Computational Linguistics.

\bibitem[{Chiang et~al.(2023)Chiang, Li, Lin, Sheng, Wu, Zhang, Zheng, Zhuang, Zhuang, Gonzalez et~al.}]{chiang2023vicuna}
Chiang, W.-L.; Li, Z.; Lin, Z.; Sheng, Y.; Wu, Z.; Zhang, H.; Zheng, L.; Zhuang, S.; Zhuang, Y.; Gonzalez, J.~E.; et~al. 2023.
\newblock Vicuna: An open-source chatbot impressing gpt-4 with 90\%* chatgpt quality.
\newblock \emph{See https://vicuna. lmsys. org (accessed 14 April 2023)}, 2(3): 6.

\bibitem[{Conover et~al.(2023)Conover, Hayes, Mathur, Xie, Wan, Shah, Ghodsi, Wendell, Zaharia, and Xin}]{conover2023free}
Conover, M.; Hayes, M.; Mathur, A.; Xie, J.; Wan, J.; Shah, S.; Ghodsi, A.; Wendell, P.; Zaharia, M.; and Xin, R. 2023.
\newblock Free dolly: Introducing the world’s first truly open instruction-tuned llm.
\newblock \emph{Company Blog of Databricks}.

\bibitem[{Cui et~al.(2024)Cui, Yuan, Ding, Yao, He, Zhu, Ni, Xie, Xie, Lin, Liu, and Sun}]{cui2024ultrafeedback}
Cui, G.; Yuan, L.; Ding, N.; Yao, G.; He, B.; Zhu, W.; Ni, Y.; Xie, G.; Xie, R.; Lin, Y.; Liu, Z.; and Sun, M. 2024.
\newblock {ULTRAFEEDBACK}: Boosting Language Models with Scaled {AI} Feedback.
\newblock In \emph{Forty-first International Conference on Machine Learning}.

\bibitem[{Ding et~al.(2023)Ding, Chen, Xu, Qin, Hu, Liu, Sun, and Zhou}]{ding-etal-2023-enhancing}
Ding, N.; Chen, Y.; Xu, B.; Qin, Y.; Hu, S.; Liu, Z.; Sun, M.; and Zhou, B. 2023.
\newblock Enhancing Chat Language Models by Scaling High-quality Instructional Conversations.
\newblock In Bouamor, H.; Pino, J.; and Bali, K., eds., \emph{Proceedings of the 2023 Conference on Empirical Methods in Natural Language Processing}, 3029--3051. Singapore: Association for Computational Linguistics.

\bibitem[{Dubey et~al.(2024)Dubey, Jauhri, Pandey, Kadian, Al-Dahle, Letman, Mathur, Schelten, Yang, Fan et~al.}]{dubey2024llama}
Dubey, A.; Jauhri, A.; Pandey, A.; Kadian, A.; Al-Dahle, A.; Letman, A.; Mathur, A.; Schelten, A.; Yang, A.; Fan, A.; et~al. 2024.
\newblock The llama 3 herd of models.
\newblock \emph{arXiv preprint arXiv:2407.21783}.

\bibitem[{Grattafiori et~al.(2024)Grattafiori, Dubey, Jauhri, Pandey, Kadian, Al-Dahle, Letman, Mathur, Schelten, Vaughan et~al.}]{grattafiori2024llama}
Grattafiori, A.; Dubey, A.; Jauhri, A.; Pandey, A.; Kadian, A.; Al-Dahle, A.; Letman, A.; Mathur, A.; Schelten, A.; Vaughan, A.; et~al. 2024.
\newblock The llama 3 herd of models.
\newblock \emph{arXiv preprint arXiv:2407.21783}.

\bibitem[{Hendrycks et~al.(2021)Hendrycks, Burns, Kadavath, Arora, Basart, Tang, Song, and Steinhardt}]{hendrycks2021measuring}
Hendrycks, D.; Burns, C.; Kadavath, S.; Arora, A.; Basart, S.; Tang, E.; Song, D.; and Steinhardt, J. 2021.
\newblock Measuring Mathematical Problem Solving With the {MATH} Dataset.
\newblock In \emph{Thirty-fifth Conference on Neural Information Processing Systems Datasets and Benchmarks Track (Round 2)}.

\bibitem[{Hu et~al.(2023)Hu, Luo, Wang, Cheng, Liu, and Sun}]{hu-etal-2023-wont}
Hu, S.; Luo, Y.; Wang, H.; Cheng, X.; Liu, Z.; and Sun, M. 2023.
\newblock Won`t Get Fooled Again: Answering Questions with False Premises.
\newblock In Rogers, A.; Boyd-Graber, J.; and Okazaki, N., eds., \emph{Proceedings of the 61st Annual Meeting of the Association for Computational Linguistics (Volume 1: Long Papers)}, 5626--5643. Toronto, Canada: Association for Computational Linguistics.

\bibitem[{Ji et~al.(2024{\natexlab{a}})Ji, Chen, Lou, Hong, Zhang, Pan, Qiu, Dai, and Yang}]{ji2024aligner}
Ji, J.; Chen, B.; Lou, H.; Hong, D.; Zhang, B.; Pan, X.; Qiu, T.; Dai, J.; and Yang, Y. 2024{\natexlab{a}}.
\newblock Aligner: Efficient Alignment by Learning to Correct.
\newblock In \emph{The Thirty-eighth Annual Conference on Neural Information Processing Systems}.

\bibitem[{Ji et~al.(2024{\natexlab{b}})Ji, Hong, Zhang, Chen, Dai, Zheng, Qiu, Zhou, Wang, Li et~al.}]{ji2024pku}
Ji, J.; Hong, D.; Zhang, B.; Chen, B.; Dai, J.; Zheng, B.; Qiu, T.; Zhou, J.; Wang, K.; Li, B.; et~al. 2024{\natexlab{b}}.
\newblock Pku-saferlhf: Towards multi-level safety alignment for llms with human preference.
\newblock \emph{arXiv preprint arXiv:2406.15513}.

\bibitem[{Khanov, Burapacheep, and Li(2024)}]{khanov2024args}
Khanov, M.; Burapacheep, J.; and Li, Y. 2024.
\newblock {ARGS}: Alignment as Reward-Guided Search.
\newblock In \emph{The Twelfth International Conference on Learning Representations}.

\bibitem[{Li et~al.(2024{\natexlab{a}})Li, Wang, Grama, and Zhang}]{li2024cascade}
Li, B.; Wang, Y.; Grama, A.; and Zhang, R. 2024{\natexlab{a}}.
\newblock Cascade reward sampling for efficient decoding-time alignment.
\newblock \emph{arXiv preprint arXiv:2406.16306}.

\bibitem[{Li et~al.(2024{\natexlab{b}})Li, Liu, Wang, Li, Lan, and Shen}]{li2024dialogue}
Li, C.; Liu, X.; Wang, Y.; Li, D.; Lan, Y.; and Shen, C. 2024{\natexlab{b}}.
\newblock Dialogue for prompting: a policy-gradient-based discrete prompt generation for few-shot learning.
\newblock In \emph{Proceedings of the AAAI Conference on Artificial Intelligence}, volume~38, 18481--18489.

\bibitem[{Li et~al.(2024{\natexlab{c}})Li, Chiang, Frick, Dunlap, Wu, Zhu, Gonzalez, and Stoica}]{li2024crowdsourced}
Li, T.; Chiang, W.-L.; Frick, E.; Dunlap, L.; Wu, T.; Zhu, B.; Gonzalez, J.~E.; and Stoica, I. 2024{\natexlab{c}}.
\newblock From Crowdsourced Data to High-Quality Benchmarks: Arena-Hard and BenchBuilder Pipeline.
\newblock \emph{arXiv preprint arXiv:2406.11939}.

\bibitem[{Li et~al.(2024{\natexlab{d}})Li, Wei, Zhao, Zhang, and Zhang}]{li2024rain}
Li, Y.; Wei, F.; Zhao, J.; Zhang, C.; and Zhang, H. 2024{\natexlab{d}}.
\newblock {RAIN}: Your Language Models Can Align Themselves without Finetuning.
\newblock In \emph{The Twelfth International Conference on Learning Representations}.

\bibitem[{Li et~al.(2023)Li, Peng, He, Galley, Gao, and Yan}]{li2023guiding}
Li, Z.; Peng, B.; He, P.; Galley, M.; Gao, J.; and Yan, X. 2023.
\newblock Guiding large language models via directional stimulus prompting.
\newblock \emph{Advances in Neural Information Processing Systems}, 36: 62630--62656.

\bibitem[{Lin et~al.(2024)Lin, Ravichander, Lu, Dziri, Sclar, Chandu, Bhagavatula, and Choi}]{lin2024the}
Lin, B.~Y.; Ravichander, A.; Lu, X.; Dziri, N.; Sclar, M.; Chandu, K.; Bhagavatula, C.; and Choi, Y. 2024.
\newblock The Unlocking Spell on Base {LLM}s: Rethinking Alignment via In-Context Learning.
\newblock In \emph{The Twelfth International Conference on Learning Representations}.

\bibitem[{Lin, Hilton, and Evans(2022)}]{lin-etal-2022-truthfulqa}
Lin, S.; Hilton, J.; and Evans, O. 2022.
\newblock {T}ruthful{QA}: Measuring How Models Mimic Human Falsehoods.
\newblock In Muresan, S.; Nakov, P.; and Villavicencio, A., eds., \emph{Proceedings of the 60th Annual Meeting of the Association for Computational Linguistics (Volume 1: Long Papers)}, 3214--3252. Dublin, Ireland: Association for Computational Linguistics.

\bibitem[{Longpre et~al.(2023)Longpre, Hou, Vu, Webson, Chung, Tay, Zhou, Le, Zoph, Wei, and Roberts}]{pmlr-v202-longpre23a}
Longpre, S.; Hou, L.; Vu, T.; Webson, A.; Chung, H.~W.; Tay, Y.; Zhou, D.; Le, Q.~V.; Zoph, B.; Wei, J.; and Roberts, A. 2023.
\newblock The Flan Collection: Designing Data and Methods for Effective Instruction Tuning.
\newblock In Krause, A.; Brunskill, E.; Cho, K.; Engelhardt, B.; Sabato, S.; and Scarlett, J., eds., \emph{Proceedings of the 40th International Conference on Machine Learning}, volume 202 of \emph{Proceedings of Machine Learning Research}, 22631--22648. PMLR.

\bibitem[{Meng, Xia, and Chen(2024)}]{meng2024simpo}
Meng, Y.; Xia, M.; and Chen, D. 2024.
\newblock Sim{PO}: Simple Preference Optimization with a Reference-Free Reward.
\newblock In \emph{The Thirty-eighth Annual Conference on Neural Information Processing Systems}.

\bibitem[{Mudgal et~al.(2024)Mudgal, Lee, Ganapathy, Li, Wang, Huang, Chen, Cheng, Collins, Strohman, Chen, Beutel, and Beirami}]{mudgal2024controlled}
Mudgal, S.; Lee, J.; Ganapathy, H.; Li, Y.; Wang, T.; Huang, Y.; Chen, Z.; Cheng, H.-T.; Collins, M.; Strohman, T.; Chen, J.; Beutel, A.; and Beirami, A. 2024.
\newblock Controlled Decoding from Language Models.
\newblock In \emph{Forty-first International Conference on Machine Learning}.

\bibitem[{Ouyang et~al.(2022)Ouyang, Wu, Jiang, Almeida, Wainwright, Mishkin, Zhang, Agarwal, Slama, Ray et~al.}]{ouyang2022training}
Ouyang, L.; Wu, J.; Jiang, X.; Almeida, D.; Wainwright, C.; Mishkin, P.; Zhang, C.; Agarwal, S.; Slama, K.; Ray, A.; et~al. 2022.
\newblock Training language models to follow instructions with human feedback.
\newblock \emph{Advances in neural information processing systems}, 35: 27730--27744.

\bibitem[{Pryzant et~al.(2023)Pryzant, Iter, Li, Lee, Zhu, and Zeng}]{pryzant2023automatic}
Pryzant, R.; Iter, D.; Li, J.; Lee, Y.~T.; Zhu, C.; and Zeng, M. 2023.
\newblock Automatic prompt optimization with" gradient descent" and beam search.
\newblock \emph{arXiv preprint arXiv:2305.03495}.

\bibitem[{Rafailov et~al.(2024)Rafailov, Sharma, Mitchell, Manning, Ermon, and Finn}]{rafailov2024direct}
Rafailov, R.; Sharma, A.; Mitchell, E.; Manning, C.~D.; Ermon, S.; and Finn, C. 2024.
\newblock Direct preference optimization: Your language model is secretly a reward model.
\newblock \emph{Advances in Neural Information Processing Systems}, 36.

\bibitem[{Shin et~al.(2020)Shin, Razeghi, Logan~IV, Wallace, and Singh}]{shin2020autoprompt}
Shin, T.; Razeghi, Y.; Logan~IV, R.~L.; Wallace, E.; and Singh, S. 2020.
\newblock Autoprompt: Eliciting knowledge from language models with automatically generated prompts.
\newblock \emph{arXiv preprint arXiv:2010.15980}.

\bibitem[{Song et~al.(2024)Song, Yu, Lang, Yu, Huang, Wang, and Li}]{song-etal-2024-scaling}
Song, F.; Yu, B.; Lang, H.; Yu, H.; Huang, F.; Wang, H.; and Li, Y. 2024.
\newblock Scaling Data Diversity for Fine-Tuning Language Models in Human Alignment.
\newblock In Calzolari, N.; Kan, M.-Y.; Hoste, V.; Lenci, A.; Sakti, S.; and Xue, N., eds., \emph{Proceedings of the 2024 Joint International Conference on Computational Linguistics, Language Resources and Evaluation (LREC-COLING 2024)}, 14358--14369. Torino, Italia: ELRA and ICCL.

\bibitem[{Wang et~al.(2024{\natexlab{a}})Wang, Xiong, Xie, Zhao, and Zhang}]{wang2024interpretable}
Wang, H.; Xiong, W.; Xie, T.; Zhao, H.; and Zhang, T. 2024{\natexlab{a}}.
\newblock Interpretable Preferences via Multi-Objective Reward Modeling and Mixture-of-Experts.
\newblock \emph{arXiv preprint arXiv:2406.12845}.

\bibitem[{Wang et~al.(2024{\natexlab{b}})Wang, Liu, Liang, Huang, Wang, Yang, Shen, Feng, Zhang, Guan et~al.}]{wang2024minstrel}
Wang, M.; Liu, Y.; Liang, X.; Huang, Y.; Wang, D.; Yang, X.; Shen, S.; Feng, S.; Zhang, X.; Guan, C.; et~al. 2024{\natexlab{b}}.
\newblock Minstrel: Structural Prompt Generation with Multi-Agents Coordination for Non-AI Experts.
\newblock \emph{arXiv preprint arXiv:2409.13449}.

\bibitem[{Wang et~al.(2023{\natexlab{a}})Wang, Li, Chen, Cai, Zhu, Lin, Cao, Liu, Liu, and Sui}]{wang2023large}
Wang, P.; Li, L.; Chen, L.; Cai, Z.; Zhu, D.; Lin, B.; Cao, Y.; Liu, Q.; Liu, T.; and Sui, Z. 2023{\natexlab{a}}.
\newblock Large language models are not fair evaluators.
\newblock \emph{arXiv preprint arXiv:2305.17926}.

\bibitem[{Wang et~al.(2023{\natexlab{b}})Wang, Li, Wang, Bai, Luo, Zhang, Jojic, Xing, and Hu}]{wang2023promptagent}
Wang, X.; Li, C.; Wang, Z.; Bai, F.; Luo, H.; Zhang, J.; Jojic, N.; Xing, E.~P.; and Hu, Z. 2023{\natexlab{b}}.
\newblock Promptagent: Strategic planning with language models enables expert-level prompt optimization.
\newblock \emph{arXiv preprint arXiv:2310.16427}.

\bibitem[{Wang et~al.(2023{\natexlab{c}})Wang, Kordi, Mishra, Liu, Smith, Khashabi, and Hajishirzi}]{wang-etal-2023-self-instruct}
Wang, Y.; Kordi, Y.; Mishra, S.; Liu, A.; Smith, N.~A.; Khashabi, D.; and Hajishirzi, H. 2023{\natexlab{c}}.
\newblock Self-Instruct: Aligning Language Models with Self-Generated Instructions.
\newblock In Rogers, A.; Boyd-Graber, J.; and Okazaki, N., eds., \emph{Proceedings of the 61st Annual Meeting of the Association for Computational Linguistics (Volume 1: Long Papers)}, 13484--13508. Toronto, Canada: Association for Computational Linguistics.

\bibitem[{Yue et~al.(2024)Yue, Qu, Zhang, Fu, Huang, Sun, Su, and Chen}]{yue2024mammoth}
Yue, X.; Qu, X.; Zhang, G.; Fu, Y.; Huang, W.; Sun, H.; Su, Y.; and Chen, W. 2024.
\newblock {MA}mmo{TH}: Building Math Generalist Models through Hybrid Instruction Tuning.
\newblock In \emph{The Twelfth International Conference on Learning Representations}.

\bibitem[{Zheng et~al.(2024)Zheng, Zhang, Zhang, Ye, and Luo}]{zheng-etal-2024-llamafactory}
Zheng, Y.; Zhang, R.; Zhang, J.; Ye, Y.; and Luo, Z. 2024.
\newblock {L}lama{F}actory: Unified Efficient Fine-Tuning of 100+ Language Models.
\newblock In Cao, Y.; Feng, Y.; and Xiong, D., eds., \emph{Proceedings of the 62nd Annual Meeting of the Association for Computational Linguistics (Volume 3: System Demonstrations)}, 400--410. Bangkok, Thailand: Association for Computational Linguistics.

\bibitem[{Zhou et~al.(2022)Zhou, Muresanu, Han, Paster, Pitis, Chan, and Ba}]{zhou2022large}
Zhou, Y.; Muresanu, A.~I.; Han, Z.; Paster, K.; Pitis, S.; Chan, H.; and Ba, J. 2022.
\newblock Large language models are human-level prompt engineers.
\newblock In \emph{The eleventh international conference on learning representations}.

\end{thebibliography}
\end{document}